\documentclass[letterpaper, 10 pt, conference]{ieeeconf}  

\IEEEoverridecommandlockouts                              

\usepackage{color}
\usepackage{algorithm}
\usepackage{algpseudocode}
\usepackage{bm}
\usepackage{graphicx}
\usepackage{amsmath}
\usepackage{array,multirow}
\usepackage[style=ieee]{biblatex}
\usepackage{siunitx}
\usepackage{subcaption}

\title{\LARGE \bf
Multi-Objective Sparse Sensing with Ergodic Optimization}
\author{Ananya Rao$^{1}$ and Howie Choset$^{1}$
\thanks{$^{1}$Robotics Institute, Carnegie Mellon University, Pittsburgh, PA, USA}%
}

\begin{document}

\maketitle
\thispagestyle{empty}
\pagestyle{empty}

\begin{abstract}
We consider a search problem where a robot has one or more types of sensors, each suited to detecting different types of targets or target information.
Often, information in the form of a distribution of possible target locations, or locations of interest, may be available to guide the search.
When multiple types of information exist, then a distribution for each type of information must also exist, thereby making the search problem that uses these distributions to guide the search a multi-objective one. 
In this paper, we consider a multi-objective search problem when the "cost" to use a sensor is limited.
To this end, we leverage the ergodic metric, which drives agents to spend time in regions proportional to the expected amount of information there.
We define the multi-objective sparse sensing ergodic (MO-SS-E) metric in order to optimize when and where each sensor measurement should be taken while planning trajectories that balance the multiple objectives.
We observe that our approach maintains coverage performance as the number of samples taken considerably degrades. 
Further empirical results on different multi-agent problem setups demonstrate the applicability of our approach for both homogeneous and heterogeneous multi-agent teams.

\end{abstract}

\section{INTRODUCTION}

In several applications, robots need to explore a target region and gather information about it, typically to cover the region or to locate objects of interest. 
Many such search and coverage applications require balancing multiple objectives. 
For example, in planetary exploration the goal is to maximize the amount of scientific information retrieved by a planetary rover, while avoiding risky areas.
Many of these objectives can be encoded as information distributions called objective maps.
Different objective maps could require different types of sensors to read in information, leading to robots having to consider multiple objectives simultaneously while coordinating different sensors.
In applications where sensing measurements are expensive, difficult, or limited in number, optimizing the use of available sensors while effectively gathering information about the region is important. 


In this work, we embrace an information-based search strategy, called ergodic search, to address the multi-objective and limited sensing budget issues. 
Trajectories planned using ergodic optimization drive the search agent to spend time in regions of the search domain in proportion to the expected amount of information there, thus balancing exploration and exploitation~\cite{mathew2011metrics}. 
Optimizing the ergodic metric ties to better coverage of the \textit{a priori} information distribution.
Ergodic search processes can be leveraged to tackle both aspects of the multi-objective sparse sensing problem.
Ergodic search processes can be augmented to incorporate sampling decision times that are solutions to a convex optimization problem, where the sampling solutions result in improved coverage of the search region~\cite{rao}.
Further, ergodic search processes also lend themselves to multi-objective planning~\cite{ren}.

\begin{figure}
\centering
    \includegraphics[width=0.75\linewidth]{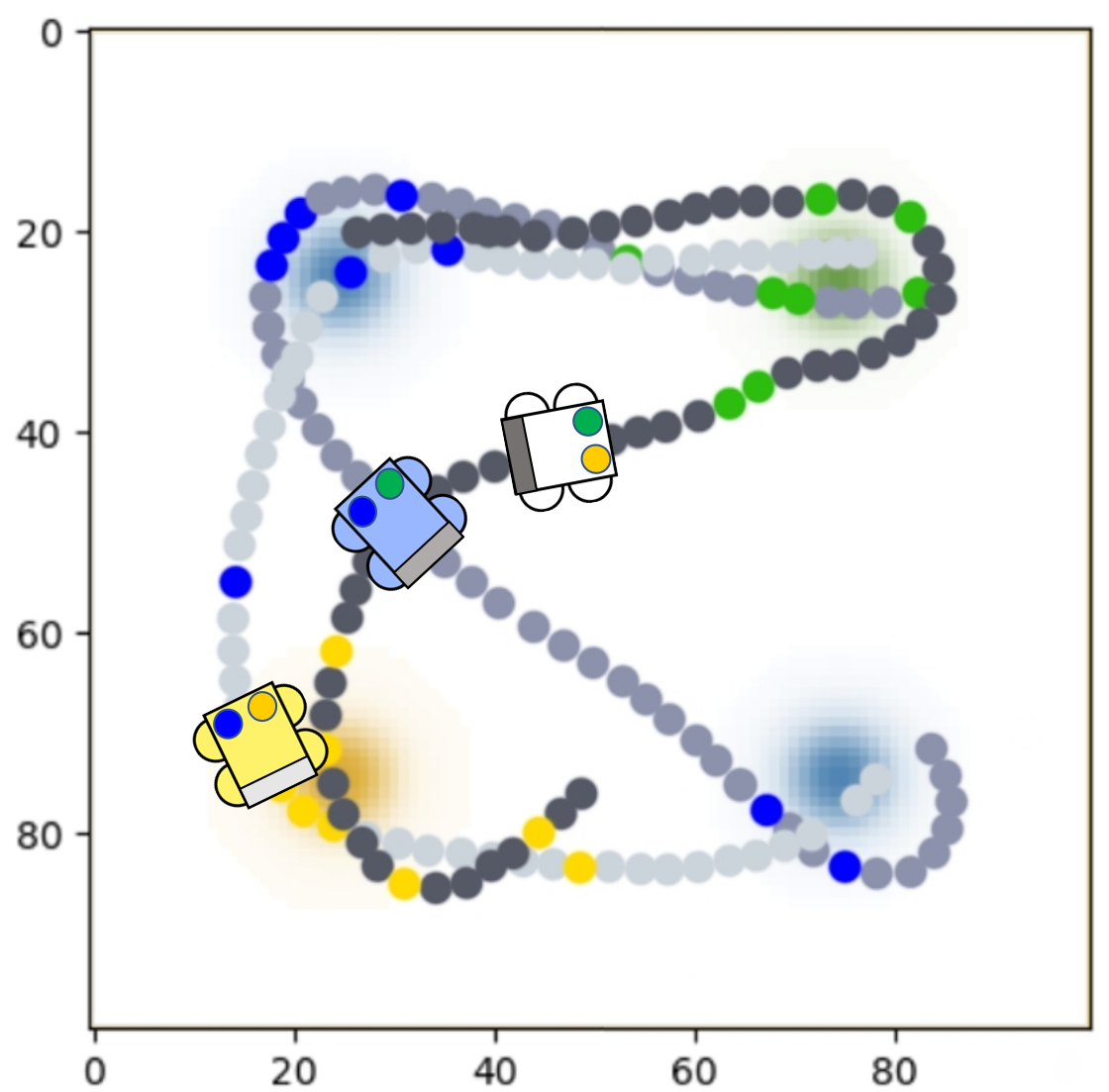}
    \caption{A heterogeneous multi-agent team covers multiple objectives using sparse sensor measurements. The agent trajectories are shown in gray on a scalarized combination of three objective maps - blue, yellow, and green. Sensing locations are indicated using the corresponding sensor's color. Each agent is equipped with a different set of sensors, indicated by the colored circles within each agent, where each sensor can read information from its corresponding objective map. The color of the agent corresponds to the sensor currently being used to take a measurement. Agents are white when they are not taking any measurements. }
    \label{fig:coverimg}
    \vspace{-0.5cm}
\end{figure}

This paper presents an approach to planning resource-efficient trajectories that balance multiple objectives for robots with multiple sensors (example result shown in Fig~\ref{fig:coverimg}). 
We introduce a metric that takes into account multiple objectives and sensing decisions for multiple sensors. 
Using this metric in ergodic optimization, we experimentally show improved coverage of multiple objective maps while requiring fewer sensing measurements.
We demonstrate the efficacy of this approach both on synthetic and real-world data, for both single and multi-agent coverage scenarios. 

\section{BACKGROUND AND PRIOR WORK}
\label{sec:background}

\subsection{Multi-Objective Planning}
A popular existing approach to multi-objective trajectory planning is genetic algorithms.
These require generating many full solutions, which can be time- and memory-intensive~\cite{fonseca1995, fonseca1999, moradi, yang}. 
Graph-based approaches tend to utilize Dijkstra's algorithm, A*, or D* search~\cite{lavin_astar, gautam, lavin_dstar}.
Such methods are effective for applications that have well-defined start and goal locations, but are not well-suited to exploration-focused problems, since they don't focus on covering a region. 

Recent work has focused on ergodic planning over multiple objective maps. 
The Multi-Objective Ergodic Search (MO-ES) method finds a set of Pareto-optimal trajectories given multiple objective maps~\cite{ren}.
The Pareto-optimal set of trajectories is built by employing single-objective ergodic search on scalarized combinations of the multiple objectives. 
The scalarized combinations are defined by a set of ``weight vectors", one for each Pareto-optimal trajectory, which describes how the objective maps were weighted and added together to form a ``scalarized" objective map that was used to plan the corresponding trajectory.

\subsection{Sparse Sensing in Ergodic Optimization}
Sparse sensing techniques are useful in applications plagued by resource limitations. 
Most prior work focuses on using sparse sensor measurements (and therefore sparse data) to accomplish tasks like localization and SLAM~\cite{intro1, miller2015ergodic} and depth reconstruction~\cite{intro2}. 
However, post-optimizing for sparse data points does not help reduce costs for limited onboard resources. 
While intelligently using limited data does help improve the performance of resource-limited robotic systems, further improvements can be made by deciding when and where to take these limited measurements.
This can be done by leveraging ergodic optimization to determine where to take the most informative measurements.

The proportion of time a robot spends at a state $x \in \mathcal{X}$, where $\mathcal{X} \subset {\rm I\!R^d}$ is the $d$-dimensional search domain is called the spatial time-average statistic of the trajectory ($\gamma$), and is defined as
\vspace{-3 mm}
\begin{equation}
C^t(x,\gamma(t))=\frac{1}{t}\int_{0}^{t} \delta(x-\gamma(\tau)) d\tau ,
\end{equation}
where the Dirac delta function is denoted as $\delta$.

The time-averaged statistics of a robot's trajectory should match the expected information density across the map. The difference between these two distributions is computed using the Fourier decomposition of each. The weighted sum of the difference between the distributions' Fourier coefficients is called the ergodic metric~\cite{mathew2011metrics}, $\Phi (\cdot)$, and is defined as
\vspace{-3 mm}
\begin{equation}
\Phi(\gamma(t))=\sum_{k=0}^{K} \alpha_k \left| c_k (\gamma(t)) - \xi_k \right|^2 ,
\label{eq:ergodic_metric}
\vspace{-2 mm}
\end{equation}
where $K$ is the number of Fourier bases chosen, $c_k$ and $\xi_k$ are the Fourier coefficients of the time-average statistics of the trajectory and the objective map being covered respectively, and $\alpha_k$ are the weights of each coefficient difference.

The required set of sensor measurements can be found using sparse ergodic optimization~\cite{rao}.
The sparse ergodic optimization problem is posed as follows,
\vspace{-0.1cm}
\begin{equation}
\begin{array}{lcl}
\label{ICAPS-eq:sparseOpt}
\mathbf{u}^*(t), \mathbf{\lambda}^*(t) & = & \operatorname*{arg\,min}_{\mathbf{u}, \mathbf{\lambda}} \Phi'(\gamma(t)),\\[0.3cm]
\mbox{subject to } \dot{\mathbf{q}} & = & f(\mathbf{q}(t),\mathbf{u}(t)),\hspace{0.2cm} \left\|\mathbf{u}(t)\right\| \leq u_{max} 
\end{array}
\vspace{-0.1cm}
\end{equation}

\noindent where $\mathbf{q}\in \mathcal{Q}$ is the state, $\mathbf{u}\in \mathcal{U}$ denotes the set of controls, and $\lambda(t) \in \{ 0,1 \}$.
$\lambda(t)$ represents the decision variable for choosing whether to take a sensor measurement or not at a given location in the search domain.
Sparsity is promoted in the sample measurements by regularizing $\lambda$ with an $L^1$ optimization~\cite{schmidt2007learning}.
The sparse ergodic metric, $\Phi_{\text{sparse}}(\cdot)$ is, 
\vspace{-0.1cm}
\begin{equation}
    \Phi_{\text{sparse}}(\gamma(t))=\sum_{k=0}^{K} \alpha_k \left| c_k (\gamma(t), \lambda(t)) - \xi_k \right|^2 + \sum|\lambda_k|.
    \label{ICAPS2023-sparseErgodicMetric}
    \vspace{-0.1cm}
\end{equation}


The spatial time-average statistics of the agent's trajectory for the sparse ergodic optimization problem are,
\vspace{-0.1cm}
\begin{equation}
    C_{\text{sparse}}^t(\mathbf{x},\gamma(t))=\frac{1}{\sum_{t} \lambda(t)}\sum_{\tau=0}^{t}  \lambda(t) \delta(\mathbf{x}-\gamma(\tau)),
\label{ICAPS2023-sparseErgodicTrajectory}
\vspace{-0.1cm}
\end{equation}

\noindent where $\lambda(t) \in \{ 0,1 \}$.

Defining $\lambda(t)$ to be an integer results in Eq~\ref{ICAPS-eq:sparseOpt} being a mixed integer programming problem.
Due to a lack of gradient information from the integer variables, such mixed integer programming problems are computationally very expensive to solve~\cite{wolsey2007mixed}. 
The sparse ergodic optimization problem can be relaxed by defining $\lambda(t)$ to be a bounded continuous variable $\lambda(t) \in [0,1]$, and projecting the resulting continuous values to the nearest integer values while adhering to the sensing budget. 
This relaxation has been experimentally shown to significantly reduce computation time without sacrificing performance of the planned trajectories~\cite{rao}.




\begin{figure}
    \centering
    \begin{subfigure}[b]{0.23\textwidth}
    \includegraphics[width=\linewidth]{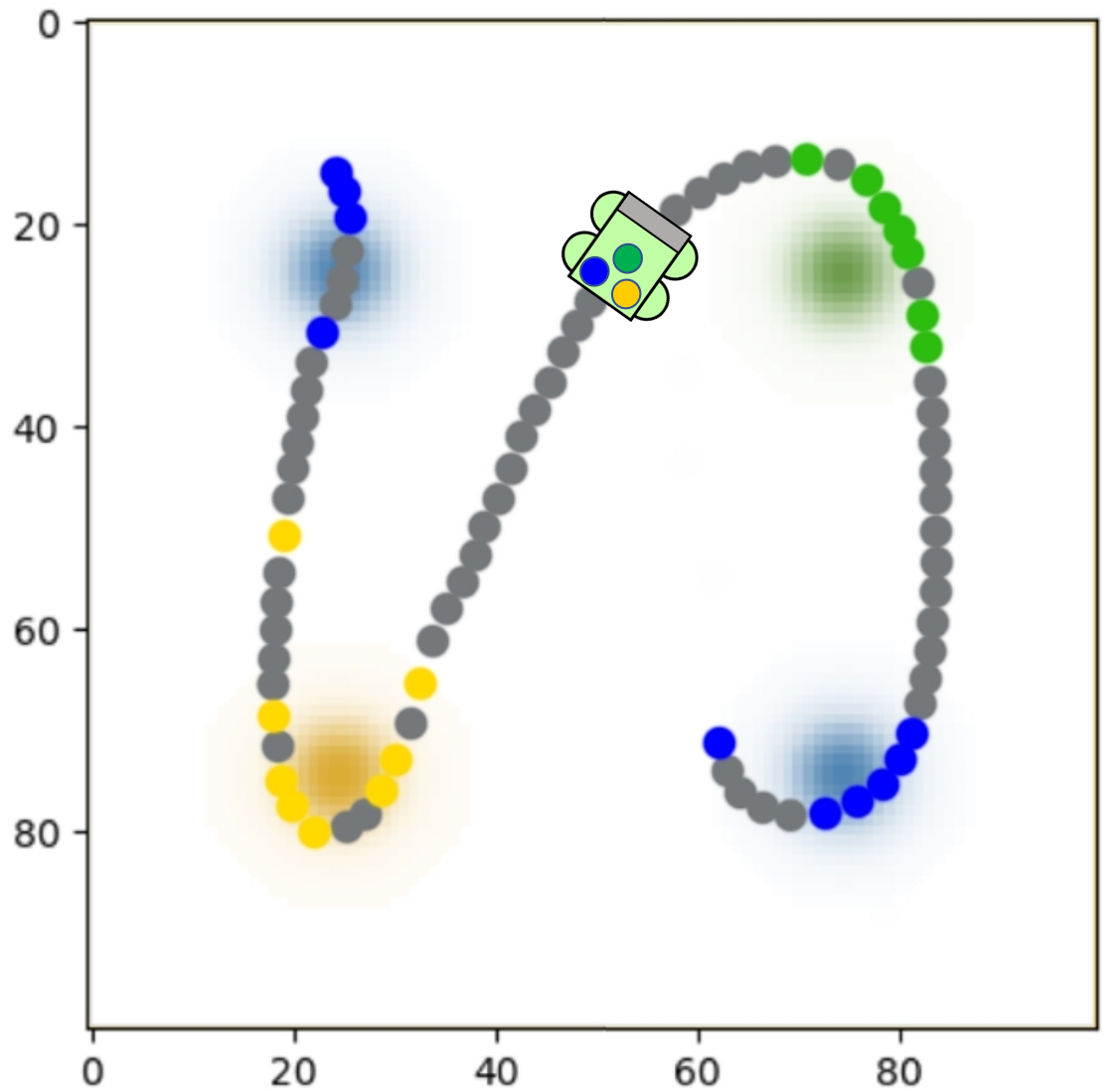}
    \end{subfigure}
    \begin{subfigure}[b]{0.23\textwidth}
    \includegraphics[width=\linewidth]{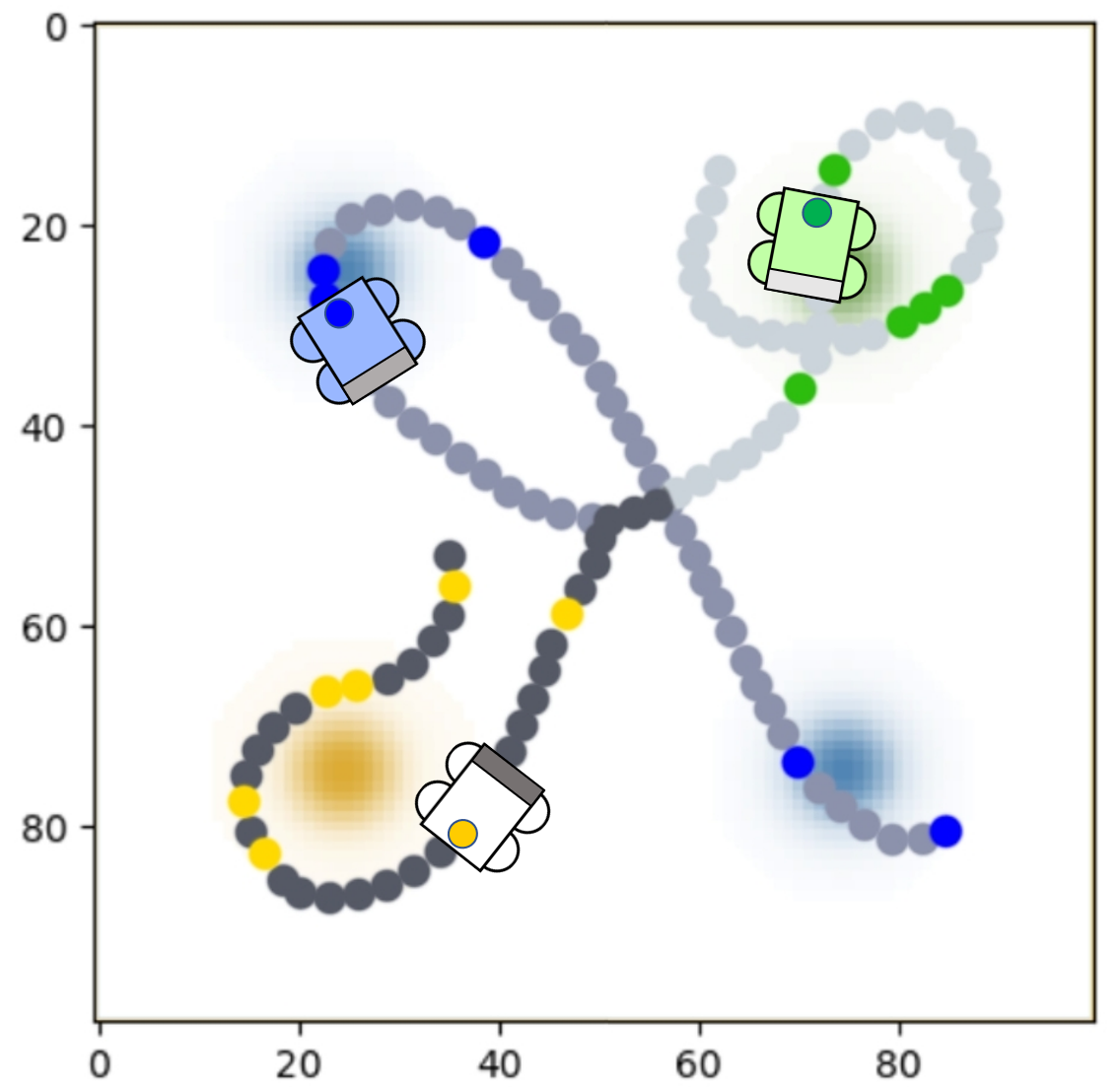}
    \end{subfigure}
    \caption{A single agent (left) and a heterogeneous multi-agent team (right) cover multiple objectives using sparse sensor measurements. Agent trajectories are shown on a scalarized combination of three objective maps - blue, yellow, and green. The colored circles on each agent represent the sensors onboard each agent. The color of the agent corresponds to the sensor currently being used to take a measurement. Agents are white when they are not taking any measurements. }
    \label{fig:single_agent}
    \vspace{-0.5cm}
\end{figure}

\section{Multi-Objective Sparse Ergodic Optimization}
\label{sec:approach}
\subsection{Multi-Objective Sparse Sensing Ergodic Metric}
This work approaches multi-objective sparse sensing in two steps.
First, we extend the ergodic metric to define the multi-objective sparse sensing ergodic metric, which takes into account multiple objective maps, and sampling decisions for different sensors.
Second, we apply this formulation to three different coverage problem variations: single agent exploration, exploration using a homogeneous multi-agent team, where each agent is equipped with the same sensor suite, and exploration using multiple heterogeneous agents, where each agent is equipped with a different sensor suite. 

Our approach to multi-objective sparse ergodic optimization is based on crafting a metric that drives trajectories to balance the multiple objectives, while jointly optimizing the trajectory and sampling decision variables associated with different sensors. 
To this end, we propose a multi-objective sparse sensing variant of the ergodic metric, that we call the \textit{MO-SS-E metric}, by defining two parts to the multi-objective sparse sensing ergodic optimization problem. 
First, the ergodic value as defined in Eq.~\ref{eq:ergodic_metric} of the agent's trajectory is evaluated on a weighted combination of the given objectives. 
Second, sparse ergodic optimization is extended to incorporate sampling decision vectors for multiple sensors. 
Note that each sensor in consideration is associated with one of the given objectives.
Further, in this work we assume that an agent can use only one sensor at a time.
An example trajectory for single agent optimization using the MO-SS-E metric is shown in Fig~\ref{fig:single_agent}.

We construct the MO-SS-E metric by augmenting the ergodic metric in the following manner
\begin{equation}
 \begin{split}
    \Phi_{\text{MO-SS-E}}(\gamma(t))= \overbrace{\sum_{i=1}^{N} \sum_{k=0}^{K} \alpha_k \left| c_k (\gamma(t), \lambda_i(t)) - \xi_{i,k} \right|^2}^{\text{Trajectory and sensing decisions for each objective map}} \\+ \underbrace{\sum_{i=1}^{N} \sum_{k=0}^{K}|\lambda_{i,k}|}_{\text{Promote sample sparsity}} + \underbrace{\sum_{k=0}^{K} \alpha_k \left| c_k (\gamma(t)) - \xi^{'}_{k} \right|^{2}}_{\text{Trajectory on combined objective maps}}
 \end{split}
 \label{ICAPS-mosseMetric}
\end{equation}

\noindent where $c_k$ and $\xi_k$ are the Fourier coefficients of the time-average statistics of an agent's trajectory $\gamma (t)$ and the $i^\text{th}$ desired spatial distribution of $N$ given objective spatial distributions, and $\alpha_k$ are the weights of each coefficient difference. $\xi^{'}_{k}$ is the Fourier coefficients of the $N$ objective distributions combined into one. 

The spatial time-average statistics of the agent's trajectory are also modified to be,
\vspace{-0.1cm}
\begin{equation}
    C_{\text{MO-SS-E}}^t(\mathbf{x},\gamma(t))=\frac{1}{\sum_{i=1}^{N}\sum_{t} \lambda_i(t)}\sum_{i=1}^{N}\sum_{\tau=0}^{t}  \lambda_i(t) \delta(\mathbf{x}-\gamma(\tau)),
\label{ICAPS-mosseTrajectory}
\vspace{-0.1cm}
\end{equation}

\noindent where $\lambda_i(t) \in \{ 0,1 \} \forall i \in [0,N) $.

We optimize for agent trajectory using the MO-SS-E metric by posing the following optimization problem, 
\vspace{-0.1cm}
\begin{equation}
\begin{array}{lcl}
\label{ICAPS2023-eq:mosseOpt}
\mathbf{u}^*(t), \mathbf{\Lambda}^*(t) & = & \operatorname*{arg\,min}_{\mathbf{u}, \mathbf{\Lambda}} \Phi'(\gamma(t)),\\[0.3cm]
\mbox{subject to } \dot{\mathbf{q}} & = & f(\mathbf{q}(t),\mathbf{u}(t)),\hspace{0.2cm} \left\|\mathbf{u}(t)\right\| \leq u_{max} 
\end{array}
\end{equation}

\noindent where $\mathbf{q}\in \mathcal{Q}$ is the state, $\mathbf{u}\in \mathcal{U}$ denotes the set of controls, and $\{\lambda_0(t), ... \lambda_{N-1}(t)\} \in \Lambda$ where $\lambda_i(t) \in \{ 0,1 \} \forall i \in [0,N)$.
$\lambda_i(t)$ represents the decision variable for sampling with the $i^\text{th}$ sensor at a given location in the search domain.
We promote sparsity in samples by regularizing $\lambda_i \forall i \in [0,N)$ with an $L^1$ optimization~\cite{schmidt2007learning}.

We leverage the MO-ES approach~\cite{ren} to combine the given objectives in order to derive $\xi_{k}^{'}$. 
MO-ES gives a set of pareto-optimal weighting schemes to combine a given set of objective maps into one information distribution using a weighted sum. 
In order to maintain autonomy in our approach, and avoid the need for a human making the choice, we use the TOPSIS method~\cite{hwang1981} to select a weighting scheme from the pareto set, which has been shown to be a good choice for this purpose~\cite{rao_icra}.

$\lambda_i(t)$ is defined to be an integer (i.e. $\lambda_i(t) \in \{0,1\}$), resulting in Eq~\ref{ICAPS2023-eq:mosseOpt} being a mixed integer programming problem.
As explained in Sec~\ref{sec:background}, we employ a relaxation of the problem Eq~\ref{ICAPS2023-eq:mosseOpt} by defining $\lambda_i(t)$ to be a bounded continuous variable $\lambda(t) \in [0,1]$, and projecting $\lambda_i$ from the continuous domain to the nearest integer value after optimization, while adhering to the sensing budget.


\subsection{Multi-Objective Coverage Problem Variants}
\begin{figure}
    \centering
    \includegraphics[width=0.85\linewidth]{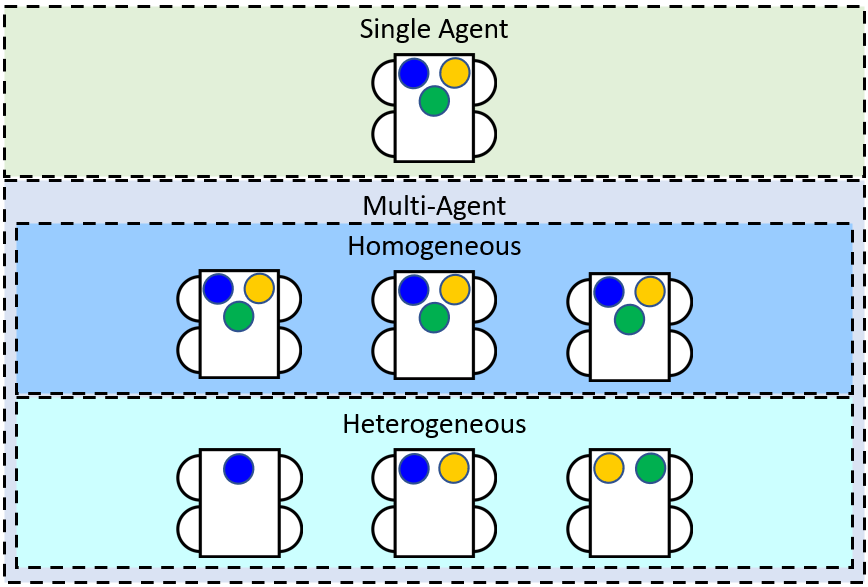}
    \caption{The different agent setups being considered in this work, with sensor suites defined assuming three objective maps - blue, yellow, and green.}
    \label{fig:scenarios}
    \vspace{-0.5cm}
\end{figure}

We consider multi-objective coverage problems in three categories (depicted in Fig~\ref{fig:scenarios}): single-agent exploration, homogeneous multi-agent exploration, and heterogeneous multi-agent exploration. 
In both the single agent and the homogeneous multi-agent cases, each agent is equipped with all of the required sensors. 
In the heterogeneous multi-agent case, each agent is equipped with a subset of sensors.

In single agent coverage problems, the agent is equipped with multiple sensors, one for each of the objective maps in consideration. 
For example, if we consider blue, green, and yellow objective maps, as in Fig~\ref{fig:single_agent}, the search agent has a blue sensor, a green sensor, and a yellow sensor on board. 
The presented formulation of the MO-SS-E metric (Eq~\ref{ICAPS-mosseMetric}) is directly applied to the single agent case. 

For a multi-agent team covering a set of objective maps, the limited number of measurements can be distributed among the different agents and different sensors. 
For a team of $M$ agents, the modified joint spatial time-average statistics of the set of agent trajectories $\{\gamma_i\}_{i=1}^M$ are defined as
\begin{equation}
\begin{split}
    C'^t(\mathbf{x},\gamma(t))=&\frac{1}{Mt\sum_{i=0}^{M-1}\sum_{i=0}^{N-1}\sum_{t} \lambda_{m,i}(t)} \\&\sum_{i=0}^{M-1}\sum_{i=0}^{N-1} \sum_{0}^{t}  \lambda_{m,i}(t) \delta(\mathbf{x}-\gamma_m(\tau)),
\label{ICAPS2023-mosseTrajectoryMulti}
\end{split}
\end{equation}
\noindent where $\lambda_{m,i}(t) \in \{ 0,1 \} \forall$ integers $i \in [0, N), m \in [0,M)$.

In the case of a homogeneous multi-agent team, each agent is equipped with all of the sensors in consideration, so, we can directly apply the MO-SS-E metric (Eq~\eqref{ICAPS-mosseMetric}). 

For a heterogeneous multi-agent team, we account for the different sensors on-board each robot by setting the sensing decision vectors for all sensors not on-board to zero.
For example, consider a coverage problem where the number of objectives $N = 3$, and the number of agents $M = 3$. 
Let agent $m = 0$ have sensor $i = 0$, agent $m = 1$ have sensor $i = 1$, and agent $m = 2$ have sensor $i = 0$. 
In this setup, we set $\lambda_{0,1} = \Vec{0}$ and $\lambda_{0,2} = \Vec{0}$ in the MO-SS-E metric (Eq~\eqref{ICAPS-mosseMetric}). 
Similarly, we set $\lambda_{1,0} = \Vec{0}$, $\lambda_{1,2} = \Vec{0}$, $\lambda_{2,0} = \Vec{0}$, and $\lambda_{2,1} = \Vec{0}$.
The same method is used to account for agents being equipped with different subsets of sensors. 


\section{EXPERIMENTS}
\label{sec:experiments}
\begin{figure}
    \centering
    \begin{subfigure}[b]{0.15\textwidth}
         \centering
         \includegraphics[width=\textwidth]{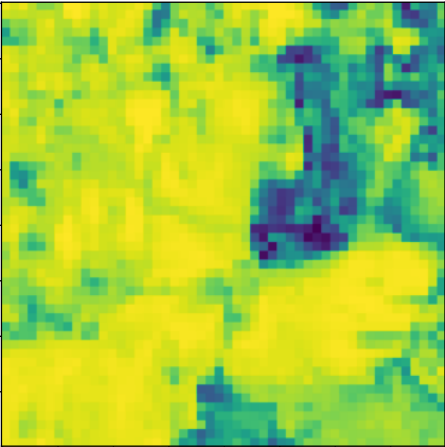}
         \caption{Entropy map}
     \end{subfigure}
     \begin{subfigure}[b]{0.15\textwidth}
         \centering
         \includegraphics[width=\textwidth]{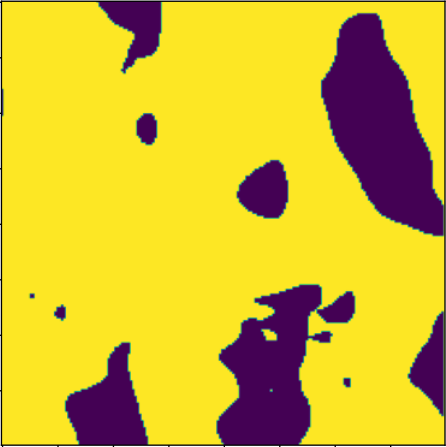}
         \caption{Shade map}
     \end{subfigure} 
     \begin{subfigure}[b]{0.15\textwidth}
         \centering
         \includegraphics[width=\textwidth]{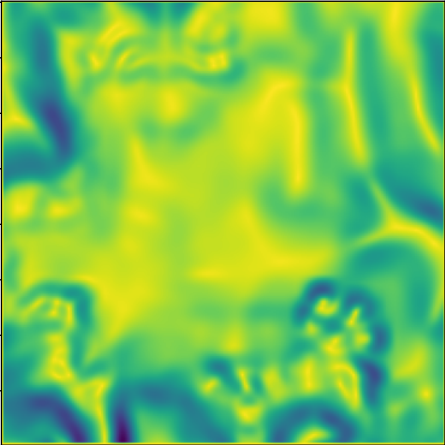}
         \caption{Slope map}
     \end{subfigure}
 \caption{Set of objective maps derived from real-world data.} 
    \label{fig:real_data}
    \vspace{-0.5cm}
\end{figure}

Our approach to multi-objective sparse sensing ergodic optimization is evaluated on two different kinds of data: synthetic Gaussian information distributions (representing objective maps in general search and coverage tasks), and objective maps from data collected of Cuprite, NV using remote sensing instruments (Fig~\ref{fig:real_data}). 
This section details how the objective maps are generated, and the experimental setups used to evaluate our approach.

\subsection{Synthetic Data}
The objective maps used for our synthetic data experiments are generated by placing Gaussian peaks at different locations to represent the entropy or uncertainty of the corresponding region, with higher values corresponding to higher uncertainty of information.

\subsection{Real-World Data}
\subsubsection{Entropy Map}
The static entropy map encodes the uncertainty of scientific information, and therefore the interest in exploring each region. 
We use the entropy map formulation proposed by Candela \textit{et al.} with the low-resolution Advanced Spaceborne Thermal Emission and Reflection Radiometer (ASTER) satellite data as the prior \cite{candela_thesis, fujisada}. We use high-resolution Airborne Visible Near Infrared Spectrometer - New Generation (AVIRIS-NG) data as a proxy for \textit{in-situ} samples \cite{green, hamlin}.
In order to focus the ergodic search on areas of high entropy, we threshold the entropy maps (setting areas of high entropy above 75\% of the max value to 1).

\subsubsection{Shade Map}
This static objective is a map of the shadowed areas of the search region, called the shade map.
The rover is solar powered, so in order to increase power generation, the rover should prefer visiting sunlit areas.
We use raycasting on a digital elevation model (DEM) of the field site to generate a map of shaded regions. 
Shadows have a low value while sunlit regions have a high value, which encourages the rover to stay in sunny areas.

\subsubsection{Slope Map}
The slope of the terrain in our test site acts as a proxy for risk, so the rover avoids high sloped areas and prefers driving over low slopes. In order to do this, the rover needs a good understanding of the slope of the regions it traverses. 
To generate this risk map, we use a Sobel image filter (used for edge detection) on a DEM of the region. 
We opt to use slope as an estimate of risk for simplicity and because of the limited information available for slip characterization \cite{candela_thesis}.

\subsection{Experiment Scenarios}
For each of the agent setups described, namely single agent, homogeneous multi-agent, and heterogeneous multi-agent, we test our approach on three-objective coverage problems using each data set.
We assume that each agent can only use one sensor at a time, based on simplified version of applications like planetary exploration. 
For the multi-agent experiments, we average results across different team sizes (three to ten agents). 
Due to numerical differences, we report the results for the two data sets separately. 

We compare our coverage results to two different baseline approaches. The first is standard ergodic optimization, where sensor measurements are uniformly distributed along the optimized trajectory. The second is a probabilistic heuristic with a two step process: first we optimize an ergodic trajectory, then we sample measurement locations based on the distribution of information under the ergodic trajectory.
Coverage performance on all objective maps is evaluated using the ergodic metric described in Eq~\ref{eq:ergodic_metric}. 
A lower ergodic metric value signals better coverage of the objective map.

The performance statistics for each method and sensing budget are averaged across 25 randomized experiment setups each, where initial information maps are varied between experiments. For each method, four different sensing budgets were used. Agents' starting positions, initial information maps, and sensing budgets are kept identical among experiments with different controllers to ensure comparable results.



\section{RESULTS AND DISCUSSION}

\bgroup
\def\arraystretch{1.15}
\begin{table*}[t]
    \centering
    \begin{tabular}{|c|
    S[table-auto-round, table-format=1.3]|
    S[table-auto-round, table-format=1.3]|
    S[table-auto-round, table-format=1.3]|
    S[table-auto-round, table-format=1.3]|
    S[table-auto-round, table-format=1.3]|
    S[table-auto-round, table-format=1.3]|
    S[table-auto-round, table-format=1.3]|
    S[table-auto-round, table-format=1.3]|
    S[table-auto-round, table-format=1.3]|
    S[table-auto-round, table-format=1.3]|
    S[table-auto-round, table-format=1.3]|
    S[table-auto-round, table-format=1.3]|
    S[table-auto-round, table-format=1.3]|
    S[table-auto-round, table-format=1.3]|
    S[table-auto-round, table-format=1.3]|}
    \multicolumn{13}{c}{Single Agent Experiments on Synthetic Data} \\
    \hline
    & \multicolumn{4}{c|}{\text{$\Phi(\gamma)$ of Objective 1}} & \multicolumn{4}{c|}{\text{$\Phi(\gamma)$ of Objective 2}} & \multicolumn{4}{c|}{\text{$\Phi(\gamma)$ of Objective 3}} \\
    \hline
    \textbf{Sensing Budget Percent} & \textbf{10} & \textbf{25} & \textbf{50} & \textbf{85} & \textbf{10} & \textbf{25} & \textbf{50} & \textbf{85} & \textbf{10} & \textbf{25}  & \textbf{50}  & \textbf{85} \\
    \hline
    Uniform Sampling
    & 0.081214 & 0.078954 & 0.0754876 & 0.07377234
    & 0.137024894 & 0.1338593 & 0.12990483 & 0.12644653
    & 0.07285939 & 0.071035784 & 0.0701248 & 0.06746046 \\
    \hline
    Probabilistic Heuristic
    & \textbf{0.075} & \textbf{0.061} & 0.05323672078 & 0.07160617451
    & 0.1146251947 & 0.09995591453 & 0.0882266605 & \textbf{0.125}
    & 0.06438089536 & 0.0612049 & 0.05697495011 & 0.06707130029 \\
    \hline
    MO-SS-E Metric
    & 0.075548686 & 0.06547183 & \textbf{0.052} & \textbf{0.071}
    & \textbf{0.114} & \textbf{0.089} & \textbf{0.087} & \textbf{0.125}
    & \textbf{0.062} & \textbf{0.058} & \textbf{0.052} & \textbf{0.064}  \\
    \hline
    \multicolumn{13}{c}{} \\
   \multicolumn{13}{c}{Single Agent Experiments on Real-World Data} \\
    \hline
    & \multicolumn{4}{c|}{\text{$\Phi(\gamma)$ of Entropy Map}} & \multicolumn{4}{c|}{\text{$\Phi(\gamma)$ of Shade Map}} & \multicolumn{4}{c|}{\text{$\Phi(\gamma)$ of Slope Map}} \\
    \hline
    \textbf{Sensing Budget Percent} & \textbf{10} & \textbf{25} & \textbf{50} & \textbf{85} & \textbf{10} & \textbf{25} & \textbf{50} & \textbf{85} & \textbf{10} & \textbf{25}  & \textbf{50}  & \textbf{85} \\
    \hline
    Uniform Sampling
    & 5.493856 & 5.109353 & 4.8029344 & 4.1204552
    & 6.928431 & 6.1092348 & 5.7183024 & 4.9942042
    & 5.463856 & 5.129353 & 4.8429344 & 4.3504552 \\
    \hline
    Probabilistic Heuristic
    & 5.198351269 & 4.35404376 & 3.814821988 & 4.122542088
    & \textbf{6.532} & \textbf{5.237} & 4.059196994 & 4.977340714
    & \textbf{5.056} & \textbf{4.300} & 3.899328236 & 4.088884868 \\
    \hline
    MO-SS-E Metric
    & \textbf{5.124} & \textbf{4.349} & \textbf{3.812} & \textbf{4.093}
    & 6.5782943 & 5.248973 & \textbf{4.013} & \textbf{4.963}
    & 5.1039845 & 4.3189723 & \textbf{3.852} & \textbf{4.043}  \\
    \hline
    \end{tabular}
    \caption{Comparative evaluation of single agent experiments using the ergodic metric ($\Phi(\gamma)$).}
    \label{tab:results_single}
    \vspace{-0.3cm}
\end{table*}
\egroup

\bgroup
\def\arraystretch{1.15}
\begin{table*}[t]
    \centering
    \begin{tabular}{|c|
    S[table-auto-round, table-format=1.3]|
    S[table-auto-round, table-format=1.3]|
    S[table-auto-round, table-format=1.3]|
    S[table-auto-round, table-format=1.3]|
    S[table-auto-round, table-format=1.3]|
    S[table-auto-round, table-format=1.3]|
    S[table-auto-round, table-format=1.3]|
    S[table-auto-round, table-format=1.3]|
    S[table-auto-round, table-format=1.3]|
    S[table-auto-round, table-format=1.3]|
    S[table-auto-round, table-format=1.3]|
    S[table-auto-round, table-format=1.3]|
    S[table-auto-round, table-format=1.3]|
    S[table-auto-round, table-format=1.3]|
    S[table-auto-round, table-format=1.3]|}
    \multicolumn{13}{c}{Homogeneous Multi-Agent Experiments on Synthetic Data} \\
    \hline
    & \multicolumn{4}{c|}{\text{$\Phi(\gamma)$ of Objective 1}} & \multicolumn{4}{c|}{\text{$\Phi(\gamma)$ of Objective 2}} & \multicolumn{4}{c|}{\text{$\Phi(\gamma)$ of Objective 3}} \\
    \hline
    \textbf{Sensing Budget Percent} & \textbf{10} & \textbf{25} & \textbf{50} & \textbf{85} & \textbf{10} & \textbf{25} & \textbf{50} & \textbf{85} & \textbf{10} & \textbf{25}  & \textbf{50}  & \textbf{85} \\
    \hline
    Uniform Sampling
    & 0.08010147335 & 0.07866284831 & 0.07683556266 & 0.07532498711
    & 1.159178522 & 1.134770999 & 1.127080305 & 1.101700681
    & 0.07445161469 & 0.06932117563 & 0.0651837953 & 0.06341715653 \\
    \hline
    Probabilistic Heuristic
    & 0.07257328756 & 0.07070804386 & 0.06641064594 & 0.07357884401
    & \textbf{1.093} & \textbf{1.036} & 0.09592747702 & 1.092748713
    & \textbf{0.065} & \textbf{0.057} & 0.05701451278 & 0.0586787184 \\
    \hline
    MO-SS-E Metric
    & \textbf{0.071} & \textbf{0.068} & \textbf{0.062} & \textbf{0.068}
    & 1.118850909 & 1.040759548 & \textbf{0.084} & \textbf{1.088}
    & 0.06695291893 & \textbf{0.057} & \textbf{0.052} & \textbf{0.058}  \\
    \hline

    \multicolumn{13}{c}{} \\
   \multicolumn{13}{c}{Homogeneous Multi-Agent Experiments on Real-World Data} \\
    \hline
    & \multicolumn{4}{c|}{\text{$\Phi(\gamma)$ of Entropy Map}} & \multicolumn{4}{c|}{\text{$\Phi(\gamma)$ of Shade Map}} & \multicolumn{4}{c|}{\text{$\Phi(\gamma)$ of Slope Map}} \\
    \hline
    \textbf{Sensing Budget Percent} & \textbf{10} & \textbf{25} & \textbf{50} & \textbf{85} & \textbf{10} & \textbf{25} & \textbf{50} & \textbf{85} & \textbf{10} & \textbf{25}  & \textbf{50}  & \textbf{85} \\
    \hline
    Uniform Sampling
    & 5.411421608 & 5.106285773 & 4.792500716 & 4.053332148
    & 6.854666143 & 6.084243285 & 5.676782069 & 5.015355186
    & 5.430899121 & 5.036131101 & 4.807735639 & 4.424806299 \\
    \hline
    Probabilistic Heuristic
    & \textbf{5.074} & \textbf{4.222} & 3.805519418 & 4.069246091
    & \textbf{6.510} & \textbf{5.207} & 4.034644833 & 4.957880489
    & \textbf{5.003} & \textbf{4.225} & 3.82972488 & 3.979272296 \\
    \hline
    MO-SS-E Metric
    & 5.076254915 & 4.251961077 & \textbf{3.796} & \textbf{4.019}
    & 6.519798342 & 5.224865076 & \textbf{3.987} & \textbf{4.914}
    & 5.013806871 & 4.249252318 & \textbf{3.781} & \textbf{3.968}  \\
    \hline
    \end{tabular}
    \caption{Comparative evaluation of homogeneous multi-agent experiments using the ergodic metric ($\Phi(\gamma)$).}
    \label{tab:results_homogeneous_multi}
    \vspace{-0.3cm}
\end{table*}
\egroup

\bgroup
\def\arraystretch{1.15}
\begin{table*}[t]
    \centering
    \begin{tabular}{|c|
    S[table-auto-round, table-format=1.3]|
    S[table-auto-round, table-format=1.3]|
    S[table-auto-round, table-format=1.3]|
    S[table-auto-round, table-format=1.3]|
    S[table-auto-round, table-format=1.3]|
    S[table-auto-round, table-format=1.3]|
    S[table-auto-round, table-format=1.3]|
    S[table-auto-round, table-format=1.3]|
    S[table-auto-round, table-format=1.3]|
    S[table-auto-round, table-format=1.3]|
    S[table-auto-round, table-format=1.3]|
    S[table-auto-round, table-format=1.3]|
    S[table-auto-round, table-format=1.3]|
    S[table-auto-round, table-format=1.3]|
    S[table-auto-round, table-format=1.3]|}
    \multicolumn{13}{c}{Heterogeneous Multi-Agent Experiments on Synthetic Data} \\
    \hline
    & \multicolumn{4}{c|}{\text{$\Phi(\gamma)$ of Objective 1}} & \multicolumn{4}{c|}{\text{$\Phi(\gamma)$ of Objective 2}} & \multicolumn{4}{c|}{\text{$\Phi(\gamma)$ of Objective 3}} \\
    \hline
    \textbf{Sensing Budget Percent} & \textbf{10} & \textbf{25} & \textbf{50} & \textbf{85} & \textbf{10} & \textbf{25} & \textbf{50} & \textbf{85} & \textbf{10} & \textbf{25}  & \textbf{50}  & \textbf{85} \\
    \hline
    Uniform Sampling
    & 0.0801996113 & 0.0739648581 & 0.07288127163 & 0.06948582169
    & 1.359929458 & 1.332943897 & 1.285049592 & 1.254177835
    & 0.07519838996 & 0.07129716416 & 0.06870871697 & 0.06611454477 \\
    \hline
    Probabilistic Heuristic
    & 0.07448424353 & 0.06423181185 & 0.06223753633 & 0.06943113024
    & 1.116394124 & 0.978991127 & 0.08962297478 & 1.113258093
    & \textbf{0.064} & \textbf{0.057} & 0.0520031642 & 0.06076479834 \\
    \hline
    MO-SS-E Metric
    & \textbf{0.071} & \textbf{0.063} & \textbf{0.059} & \textbf{0.068}
    & \textbf{1.112} & \textbf{0.978} & \textbf{0.075} & \textbf{1.099}
    & 0.06537396762 & 0.05853158318 & \textbf{0.048} & \textbf{0.059}  \\
    \hline

    \multicolumn{13}{c}{} \\
   \multicolumn{13}{c}{Heterogeneous Multi-Agent Experiments on Real-World Data} \\
    \hline
    & \multicolumn{4}{c|}{\text{$\Phi(\gamma)$ of Entropy Map}} & \multicolumn{4}{c|}{\text{$\Phi(\gamma)$ of Shade Map}} & \multicolumn{4}{c|}{\text{$\Phi(\gamma)$ of Slope Map}} \\
    \hline
    \textbf{Sensing Budget Percent} & \textbf{10} & \textbf{25} & \textbf{50} & \textbf{85} & \textbf{10} & \textbf{25} & \textbf{50} & \textbf{85} & \textbf{10} & \textbf{25}  & \textbf{50}  & \textbf{85} \\
    \hline
    Uniform Sampling
    & 5.433939139 & 4.992040194 & 4.653800329 & 4.370082891
    & 6.889201111 & 6.020842812 & 5.676036819 & 5.090077529
    & 5.381362554 & 4.955825718 & 4.724946974 & 4.480117121 \\
    \hline
    Probabilistic Heuristic
    & \textbf{5.049} & \textbf{4.228} & 3.718344045 & 4.048766412
    & \textbf{6.443} & \textbf{5.054} & 3.872915053 & 4.949222166
    & \textbf{4.973} & \textbf{4.188} & 3.837821595 & 3.966869177 \\
    \hline
    MO-SS-E Metric
    & 5.056486966 & 4.247624225 & \textbf{3.715} & \textbf{4.025}
    & 6.477902519 & 5.100787462 & \textbf{3.857} & \textbf{4.858}
    & 5.010376928 & 4.236530351 & \textbf{3.816} & \textbf{3.928}  \\
    \hline
    \end{tabular}
    \caption{Comparative evaluation of heterogeneous multi-agent experiments using the ergodic metric ($\Phi(\gamma)$).}
    \label{tab:results_heterogeneous_multi}
    \vspace{-0.5cm}
\end{table*}
\egroup


We empirically show that ergodic optimization using the MO-SS-E metric results in better performance on multiple objectives with lower sensing budgets in comparison to baseline methods. 
We demonstrate that our approach can be applied to single agent, homogeneous multi-agent, and heterogeneous multi-agent multi-objective coverage problems. 
Further, we experimentally show that our approach is effective for both simulated and real-world data. 
In this section we detail our numerical results. 

Looking at the results of the single agent experiments (detailed in Table~\ref{tab:results_single}), using the MO-SS-E metric resulted in better coverage performance compared to baseline methods, in terms of the ergodic metric, on many objective maps, as the sensing budget is reduced. 
For some of the objectives, such as the shade map and the slope map, the MO-SS-E metric approach shows a bigger performance degradation with sensing budget reduction. 
This could be a function of how similar the different objective maps are, since the shade and slope maps are highly correlated (as described in Sec~\ref{sec:experiments}).
We assume that an agent can only take a measurement with one sensor at a time, which negatively impacts performance on coverage problems with highly correlated objective maps, since information can be gathered from only one objective map at a time, leading to an agent needing to spend more time in any region that is interesting in multiple objectives. 

The results of multi-agent experiments (detailed in Tables~\ref{tab:results_homogeneous_multi} and~\ref{tab:results_heterogeneous_multi}) show the same comparative results: optimization with the MO-SS-E metric outperforms both taking uniform samples and using a probabilistic heuristic to distribute samples over standard ergodic trajectories as sensing budget reduces.
An example result from our heterogeneous multi-agent experiments is shown in Fig~\ref{fig:single_agent}.

We observe that the probabilistic heuristic approach leads to better coverage performance than the MO-SS-E metric approach at very low sensing budgets for multi-agent experiments run on real-world data collected at Cuprite, NV.
In addition to the performance degradation caused by each agent being able to use only one sensor at a time, the performance differences could be an artifact of the way we combine objective maps.
For example, if the chosen weighting scheme in these cases weights one objective map significantly higher than the others, optimization using the MO-SS-E metric would result in trajectories that focus on covering the higher weighted objective, leading to worse performance on the other objectives.

Our empirical results support the notion that jointly optimizing for agent trajectory and sensing locations improves performance.
Further, we see that specifically accounting for balancing multiple objectives in sparse ergodic optimization leads to better use of limited sensing resources.

\section{CONCLUSIONS}
In this paper we extend the ergodic metric to jointly optimize for coverage of multiple information distributions, sensing trajectory, and the decision of where to take sensing measurements with different sensors. 
We define the multi-objective sparse sensing ergodic metric in which the decision to take a measurement with each sensor is encoded in a vector of decision variables, and the sensing trajectory is additionally optimized over a pareto-efficient combination of the objective maps.
We further explore the application of the MO-SS-E metric to multi-objective coverage problems using a single agent, homogeneous multi-agent teams, and heterogeneous multi-agent teams. 

This work experimentally shows that optimizing trajectories with the MO-SS-E metric leads to better coverage performance with smaller numbers of samples. 
Numerical results on both simulated Gaussian information maps and real-world data show that ergodic optimization with the MO-SS-E metric leads to better coverage performance than baseline approaches, particularly for lower sensing budgets. 
The performance of this formulation could lead to wider applicability of robotic solutions in resource-limited settings. 

Future work will look into sensor fusion and accounting for taking multiple measurements simultaneously. 
This work assumes the availability of accurate \textit{a priori} information maps, and accurate sensor measurements, which is not the case for many real-world applications. 
Future work will investigate using the MO-SS-E metric with inaccurate \textit{a priori} information, and incorporating sensor noise.







\printbibliography

\end{document}